\documentclass{article}

\usepackage[english]{babel}
\usepackage{setspace}
\usepackage[letterpaper,top=2cm,bottom=2cm,left=3cm,right=3cm,marginparwidth=1.75cm]{geometry}

\usepackage{amsmath}
\usepackage{graphicx}
\usepackage{subcaption}
\usepackage[colorlinks=true, allcolors=blue]{hyperref}

\title{New Epochs in AI Supervision: Design and Implementation of an Autonomous Radiology AI Monitoring System}

\usepackage{authblk}

\author{Vasantha Kumar Venugopal \thanks{Corresponding author: vasanth.venugopal@carpl.ai}}
\author{Abhishek Gupta \thanks{Corresponding author: abhishek.gupta@carpl.ai}}
\author{Rohit Takhar}
\author{Vidur Mahajan}
\affil{CARPL.ai, New Delhi, India}

\begin{document}
\maketitle

\begin{abstract}

With the increasingly widespread adoption of AI in healthcare, maintaining the accuracy and reliability of AI models in clinical practice has become crucial. In this context, we introduce novel methods for monitoring the performance of radiology AI classification models in practice, addressing the challenges of obtaining real-time ground truth for performance monitoring. We propose two metrics - predictive divergence and temporal stability - to be used for preemptive alert of AI performance changes. Predictive divergence, measured using Kullback-Leibler and Jensen-Shannon divergences, evaluates model accuracy by comparing predictions with those of two supplementary models. Temporal stability is assessed through a comparison of current predictions against historical moving averages, identifying potential model decay or data drift. This approach was retrospectively validated using chest X-ray data from a single-center imaging clinic, demonstrating its effectiveness in maintaining AI model reliability. By providing continuous, real-time insights into model performance, our system ensures the safe and effective use of AI in clinical decision-making, paving the way for more robust AI integration in healthcare\end{abstract}

\section{Introduction}

The integration of artificial intelligence (AI) in healthcare has ushered in a novel era of clinical diagnostics and decision-making \cite{1Bohr2020, 2Alowais2023}. As these AI models rapidly permeate the medical imaging landscape, their performance and reliability \cite{3Balagurunathan2021, 4Feng2022, 5Taimoor2022} become critical factors in patient care. But a conundrum arises: how do we ensure the reliability of these models in real-time\cite{4Feng2022}, especially in the absence of blinded ground truth data? This paper offers a robust solution to the problem, proposing a novel approach to AI performance monitoring, demonstrated through a retrospective analysis of chest X-ray data.
The approach is centered around two primary components: Firstly, a surrogate estimation of accuracy based on divergence measures that leverage multiple models, and secondly, a temporal comparison that evaluates the consistency of a model by contrasting its current predictions with its historical prediction range.

\section{Challenges in the Real-time Monitoring of Healthcare AI}

The promise of AI in healthcare is tantalizing, especially in the realm of medical imaging, where deep learning models have shown potential in interpreting complex imaging data \cite{6Altman2017, 7Bi2019}, augmenting radiologists' abilities \cite{8Sorantin2021}, and expediting clinical decision-making \cite{9vanderSchaar2020, 10Debnath2020}. However, the real-world application of these models is not without challenges. The dynamic and non-deterministic nature of deep learning-based AI models necessitates continuous, real-time monitoring to ensure safe and effective usage \cite{4Feng2022}. Traditional monitoring approaches, primarily based on ground truth data and human intervention, fall short when applied to AI models due to the lack of readily available ground truth data in real-world settings.

\subsection{Ground Truth in the Shadows}
The absence of ground truth data poses a significant challenge in AI model monitoring. Typically, model performance is evaluated against a set of ground truth labels. However, in a clinical setting, acquiring real-time ground truth data is often impractical or impossible \cite{11soin}.

Current strategies for monitoring AI models often resort to comparing the model's predictions with final diagnostic reports generated by healthcare professionals. While this approach provides a reference for assessing model performance, it is essentially a form of retrospective monitoring and thus does not provide real-time insight\cite{11soin, 12rabanser}. Furthermore, this methodology can be influenced by the AI's predictions, leading to a potential automation bias, as clinicians might adjust their reports based on the AI's output.

In an ideal world, monitoring of AI models in healthcare would involve a parallel, blinded ground truthing process, where model predictions are independently compared against a set of outcomes not influenced by the AI model itself. This constraint necessitates the exploration of alternative monitoring strategies capable of providing valuable insights into the model's performance, independent of ground truth data.

\subsection{Model decay}

Model decay, a phenomenon where an AI model's performance degrades over time, further complicates the monitoring process. The potential causes of model decay are manifold and can be specific to the radiology context. For example, evolving clinical practices, changes in the underlying data distribution, or shifts in patient demographics can all contribute to a model's diminishing accuracy over time. 

Furthermore, changes in imaging acquisition protocols or hardware updates might introduce variations in the image characteristics, potentially affecting the performance of AI models \cite{13Rashed2023, 14Willemink2020}. As AI models are trained on specific types of data, shifts in imaging protocols can lead to discrepancies between the training data and real-world inputs, causing the model to produce less reliable results. 

Technical issues such as image artifacts, variations in patient positioning, and differences in image pre-processing can also contribute to model decay \cite{15Salvi2021}. Over time, these inconsistencies can cause the model to drift away from its initially learned patterns, affecting its overall performance.

Moreover, in the field of radiology, advancements in disease understanding and diagnostic criteria may also lead to model decay. As new knowledge emerges and diagnostic guidelines are updated, models trained on older data might start to lag in performance due to their inability to account for these changes\cite{14Willemink2020, 16SuazaMedina2023}.

Detecting and addressing model decay is therefore paramount in maintaining the reliability of AI models in clinical practice. This not only involves regular re-training and updates of the model to align with the most recent data and practices but also the development of robust monitoring systems capable of identifying early signs of model decay.

\subsection{Ensuring Accuracy and Consistency}
Accuracy and consistency are vital attributes of AI models used in clinical decision-making. Accuracy refers to the model's ability to provide correct predictions or classifications \cite{4Feng2022}, while consistency relates to the model's stability and reliability in delivering similar predictions for the same inputs. In a clinical context, while accuracy remains crucial, consistency reflects the model's ability to provide consistent outcomes for repeated evaluations of the same input, regardless of external factors or minor variations in the data. 
To ensure both accuracy and consistency, it is crucial to establish comprehensive monitoring parameters. In addition to traditional accuracy metrics, monitoring systems should incorporate metrics that specifically assess consistency. These metrics can include measures of prediction stability, such as the variance or standard deviation of predictions for a given input. In lab medicine, precision refers to the spread of repeated measurements. Accurate measurements are close to the true value, irrespective of the spread of the measurements. Precise measurements are close to each other, irrespective of their deviation from the true value. \cite{17Naphade2023}.



\begin{figure}[ht]
    \centering
    \includegraphics[width=0.8\linewidth]{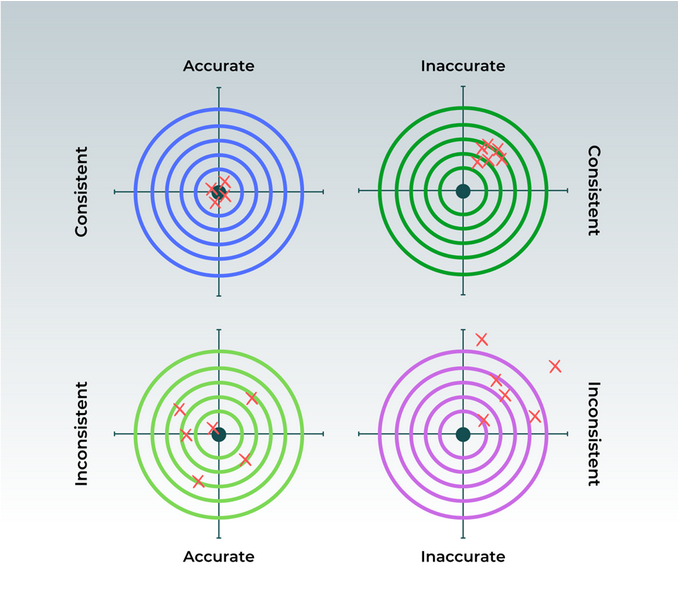}
    \caption{Accuracy vs. Consistency: Quadrants depicting the ideal and varied outcomes of AI predictions}
    \label{fig:figure6}
\end{figure}

\subsection{Balancing Real-time Monitoring with Clinical Workflow}
Integrating real-time monitoring seamlessly into the clinical workflow is essential to minimize disruptions and ensure efficient use of resources. The monitoring system should be user-friendly, providing clear and interpretable information to healthcare professionals without hindering their workflow. 



\section{Considerations for designing user-friendly monitoring systems}
To address the above-mentioned challenges and deploy a robust user-friendly intuitive model monitoring system, we propose a bifurcated monitoring process - the
first part focuses on generating alerts for potential model malfunction or failure, while the second part
investigates the reasons for such malfunction or failure. For the alert mechanism, we introduce two key metrics: Predictive Divergence and Temporal Stability. These metrics offer insights into the model's accuracy and consistency, respectively, serving as critical indicators of its performance.

\subsection{Predictive Divergence}
Predictive Divergence is in a true sense, a surrogate measure of the model's accuracy. The cornerstone of this approach is the use of multiple models that are designed for the same classification problem and measurement of divergence measures, such as the Kullback-Leibler (KL) \cite{18Ji2022} and Jensen-Shannon (JS) divergences \cite{19Menndez1997}, The model currently utilized in clinical settings is designated as the 'main model,' while the additional models are referred to as 'support models.' These metrics assess the disparity in probability distributions between the main and support models over the same time frame. A lesser divergence suggests a greater concordance between the predictions of the main model and its support counterparts, implying superior accuracy of the main model. Employing multiple support models mitigates the risk that observed divergence is attributable to the degradation of a support model, rather than the main model. The choice of support models is critical in this setup. The support models should ideally have comple-
mentary strengths and weaknesses to the main model, providing a robust and comprehensive represen-
tation of the possible prediction outcomes.
 
\subsection{Temporal Stability}
Temporal Stability, on the other hand, is a measure of our model’s consistency over time. To evaluate the precision of our main model, we compare its current prediction distribution with the moving averages of its historical prediction distributions.  We measure the divergence scores between these distributions to detect changes in the model’s pre-
diction patterns over time, which could signal model decay, data drift or other performance
issues. A model with high temporal stability demonstrates consistent prediction patterns over time, which is crucial for the reliability of the AI system. The selection of the time window for comparison (e.g., T1-x and T2-x)
is a key consideration in this part of the monitoring process. The optimal window size may vary
depending on the specific clinical setting and the characteristics of the data, and it may require some
trial and error to determine the most suitable window size.

\section{Divergence Measures}
Divergence measures like the Kullback-Leibler (KL) divergence and the Jensen-Shannon (JS) diver-
gence allow for quantification of the dissimilarities between two probability distributions
\begin{figure}[ht]
    \centering
    \begin{subfigure}[b]{0.7\linewidth}
        \centering
        \includegraphics[width=\linewidth]{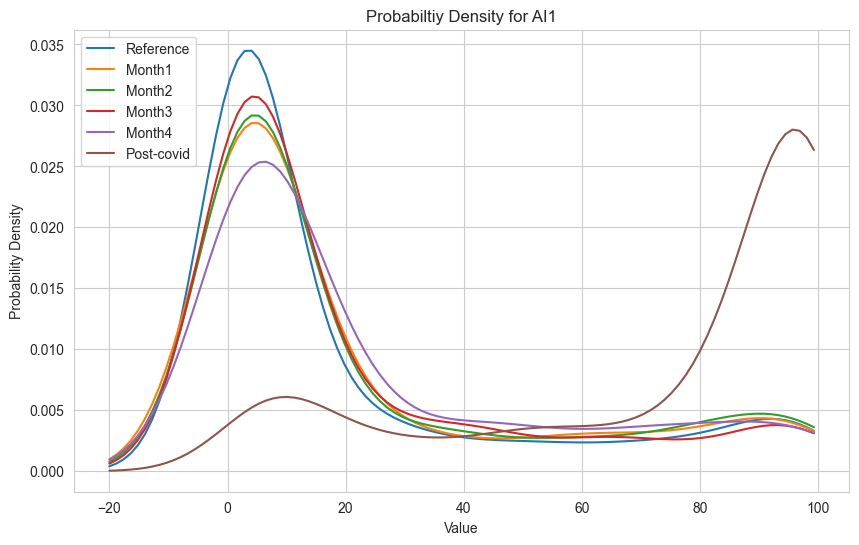}
        \label{fig:figure1}
    \end{subfigure}
    \begin{subfigure}[b]{0.45\linewidth}
        \centering
        \includegraphics[width=\linewidth]{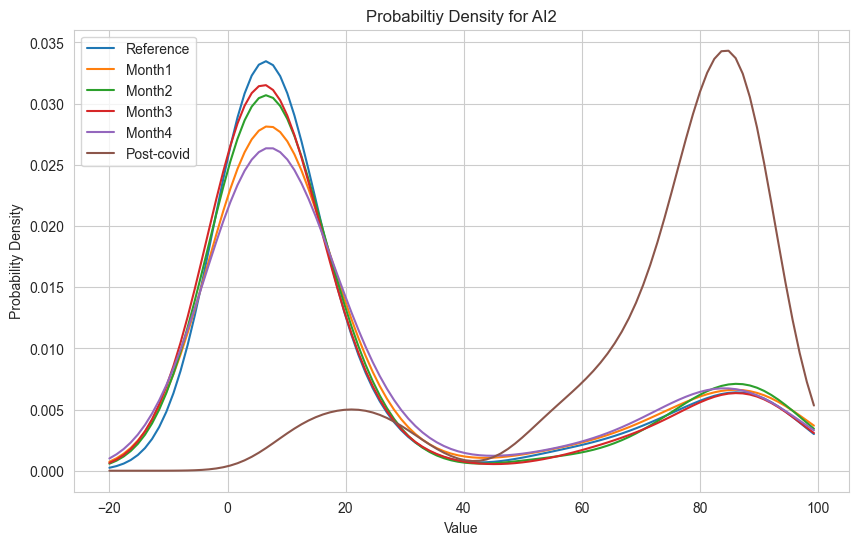}
        \label{fig:figure2}
    \end{subfigure}
    \hspace{1cm} 
    \begin{subfigure}[b]{0.45\linewidth}
        \centering
        \includegraphics[width=\linewidth]{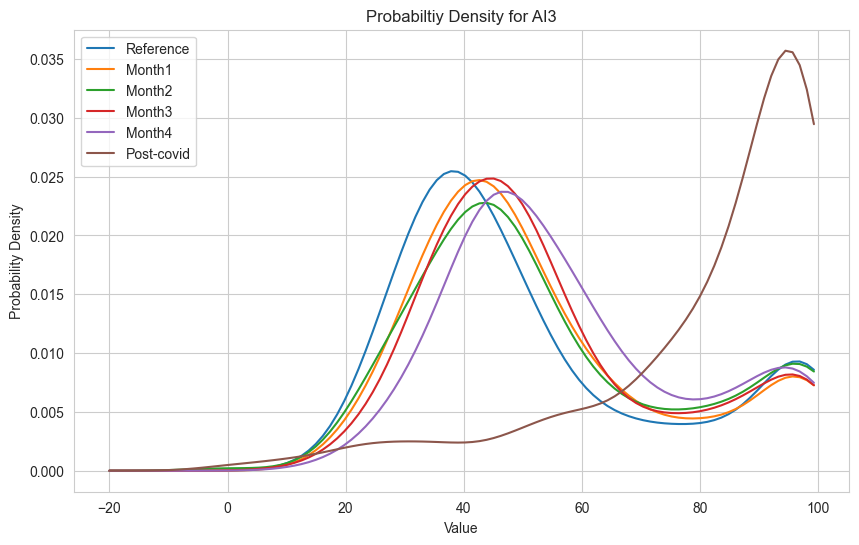}
        \label{fig:figure3}
    \end{subfigure}
    \caption{Probability Distribution for AI1 (top) AI2 (left) and AI3 (right)}
    \label{fig:figure2and3}
\end{figure}
\subsection{KL Divergence:}
Divergence measures like the Kullback-Leibler (KL) divergence and the Jensen-Shannon (JS) divergence allow for quantification of the dissimilarities between two probability distributions. 
KL divergence, short for Kullback-Leibler divergence, is a way to measure how different two probability distributions are from each other \cite{18Ji2022}. Probability distributions help us understand how likely different events are to occur. Let's say we have two probability distributions: P and Q. We want to know how much P differs from Q. The formula for KL divergence is:
\begin{equation}
\label{eq:KLDivergence}
    KL(P \| Q) = \sum (P(x) \log\left(\frac{P(x)}{Q(x)}\right))
\end{equation}

In simpler terms, KL divergence measures how much we need to change the Q distribution to match the P distribution. It gives us a number that tells us how different the two distributions are.

If KL divergence is zero, it means the distributions P and Q are exactly the same. But if KL divergence is greater than zero, it means the distributions are different from each other. It should be noted that KL divergence is not a symmetrical measure, meaning KL(P || Q) is not the same as KL(Q || P). It matters which distribution you consider as the reference (P) and which one you compare it to (Q).

\subsection{JS Divergence:}
JS divergence is a way to measure the similarity between two probability distributions \cite{19Menndez1997}. It tells us how different or similar the two distributions are. Let's say we have two distributions, P and Q, and we want to calculate the JS divergence between them.
\begin{equation}
\label{eq:JSDivergence}
    JS(P \| Q) = \frac{1}{2} \, KL(P \| M) + \frac{1}{2} \, KL(Q \| M)
\end{equation}
Where KL is the KL represents KL divergence and M represents the average distribution.
 If the JS divergence is zero, it means the distributions are very similar. The larger the JS divergence, the more different the distributions are.

JS divergence is a symmetrical measure, meaning JS(P || Q) is the same as JS(Q || P). It doesn't matter which distribution you consider as P and Q.

\section{Implementation in Action: A Case Study using Retrospective Analysis}
To demonstrate the practical applicability of our proposed system, we conducted a retrospective analysis using chest X-ray data from a single-center imaging clinic. We utilized three different commercial chest X-ray classification models, (now on referred to as AI1, AI2, and AI3) for the detection of the presence or absence of consolidation on the chest X-rays. The analysis included a total of 3,993 frontal chest X-rays. The objective was to simulate a prospective environment using this data and propose an implementation framework to evaluate AI1's accuracy and consistency through predictive divergence and temporal stability metrics.

\subsection{Study Methodology and Results:}
The study used a longitudinal retrospective study design, analyzing consecutive data sets. A reference test data set comprising of consecutive data from August till October 2019, was used as the baseline for our analysis.
Further data were collected over four continuous months—from February to May 2020. Additionally, a dataset from June 2020, a period that marked the onset of the first wave of the COVID-19 pandemic, was also included in the analysis. The data sets encompassed the following sample sizes at each time point: 
Reference Point - 969 studies,
Month 1 (February 2020)- 489 studies, 
Month 2 (March 2020)- 646 studies, Month 3 (April 2020) - 543 studies, Month 4 (May 2020)- 352 studies
Post-COVID (June 2020)- 994 studies. 
We used Jensen-Shannon Divergence (JSD) as the divergence measure to compute both predictive divergence and temporal stability metrics. The reasons for our preference for Jensen-Shannon Divergence are explained in the discussion section below. Lower JSD values indicate a close resemblance in predictions, whereas higher values signify a significant divergence. 

 \subsection{Predictive Divergence:}
\begin{figure}[ht]
    \centering
    \includegraphics[width=0.8\linewidth]{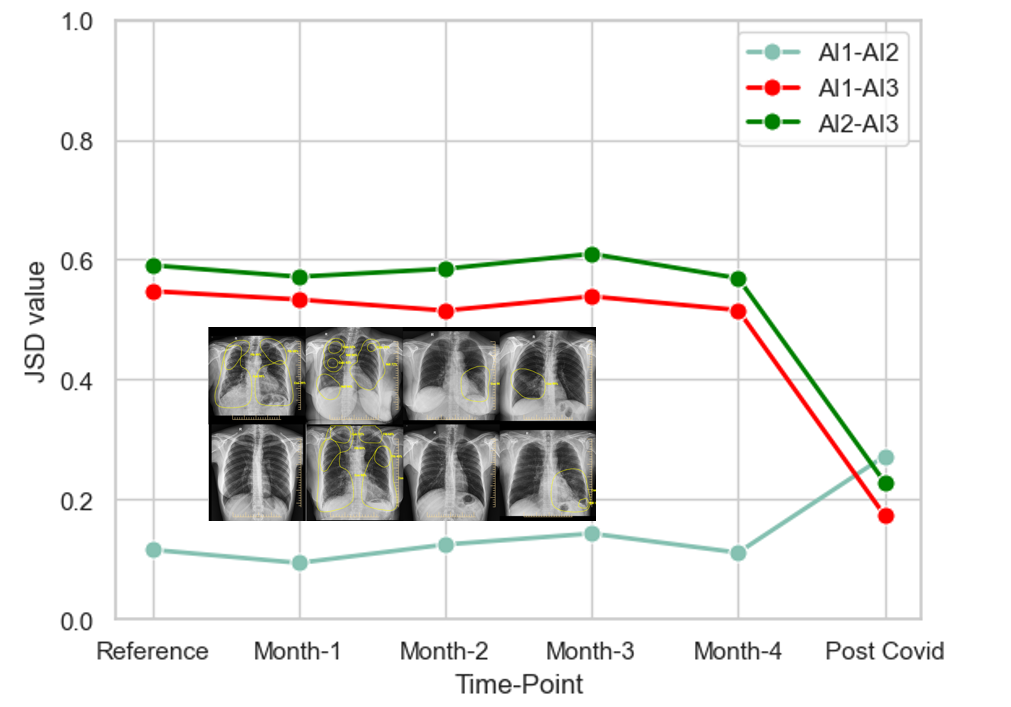}
    \caption{JSD between main model and support models}
    \label{fig:figure4}
\end{figure}

For Predictive divergence, we considered AI1 as the main model, and AI2, and AI3 as support models. For the reference data and for each month, JSD was calculated between AI1-AI2, AI1-AI3, and AI2-AI3.  The results of this analysis are provided in Table \ref{tab:JSD table} and illustrated in Figure \ref{fig:figure4}.

\begin{table}
\centering
\begin{tabular}{|c|c|c|c|c|} \hline 
Time Point & Studies & JSD (AI1-AI2) & JSD (AI1-AI3) & JSD (AI2-AI3) \\\hline
Reference Point & 969 & 0.116 & 0.547 & 0.590 \\ \hline 
Month 1 & 489 & 0.094 & 0.533 & 0.571 \\ \hline 
Month 2 & 646 & 0.124 & 0.514 & 0.584 \\ \hline 
Month 3 & 543 & 0.143 & 0.538 & 0.609 \\ \hline 
Month 4 & 352 & 0.112 & 0.515 & 0.568 \\ \hline 
Post Covid & 994 & 0.272 & 0.173 & 0.228

 \\ \hline\end{tabular}
\caption{\label{tab:JSD table}Representation of Number of studies along with JS divergence between probability distributions of the three AI solutions at each time point}
\end{table}

The results for predictive divergence, as captured by the Jensen-Shannon Divergence (JSD) values, demonstrate notable trends across different time points and between model comparisons. Initially, AI1 and AI2 show a low JSD value (0.116 at the reference point), suggesting a high degree of alignment in their predictions, which is within the acceptable tolerance range of 20 percent. As we progress through the months, we observe some fluctuations in the JSD values, with a slight increase in divergence in Month 3, indicating a potential drift in model alignment.

Post-COVID, we see a significant increase in the divergence between AI1 and AI2 to 0.272, exceeding our 20 percent tolerance threshold, which may warrant intervention. However, the divergence between AI1 and AI3, and AI2 and AI3, decreased sharply to 0.173 and 0.228, respectively. This decrease, particularly involving AI3, is intriguing and may potentially be attributed to the model's initially low specificity to consolidation, which appears to be spuriously corrected by the increased prevalence of such patterns during the COVID-19 pandemic. This hypothesis suggests that AI3's prediction performance may have aligned closer to the other models' outputs due to the changing prevalence of clinical features in the data, rather than an inherent improvement in its diagnostic specificity.

 \subsection{Temporal Stability:}
\begin{figure}[ht]
    \centering
    \includegraphics[width=0.8\linewidth]{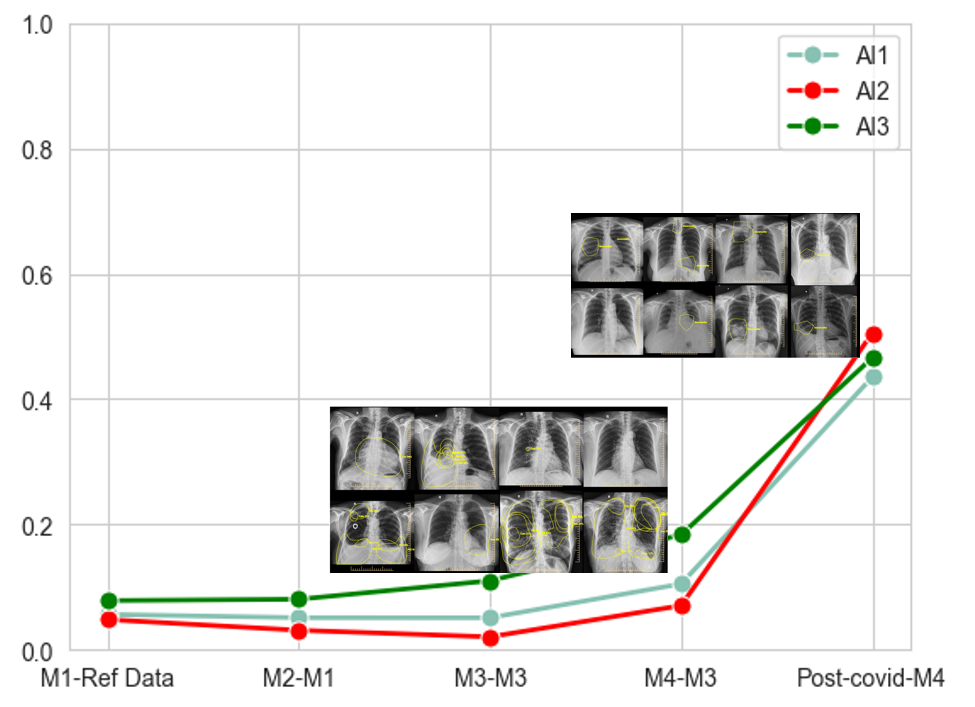}
    \caption{Temporal Stability of all three AIs at the monthly check points}
    \label{fig:figure5}
\end{figure}

For Temporal stability, we computed the Jensen-Shannon Divergence (JSD) for each AI model between the prediction distributions of consecutive months, the distributions for any given month were compared against those from the preceding month. For example, the distribution of predictions in March 2020 was measured against the distribution from February 2020, and similarly, the predictions for April 2020 were compared with those from March 2020. Given the constraints of data volume, monthly intervals were chosen as the optimal time points for this analysis. However, in a setting with more robust data flows, this model could be adapted to a daily evaluation frequency. The aim is to create a forward-moving average \cite{20Hansun} that can serve as a dynamic benchmark, allowing for the adjustment of the monitoring frequency according to the specific data characteristics and operational needs of each clinical site. The results of this analysis are provided in Table \ref{tab:JSD table 2} and illustrated in Figure \ref{fig:figure5}. 

 \begin{table}
\centering
\begin{tabular}{|c|c|c|c|c|c|} \hline 
AI Algorithm & M1-Reference Data & M2-M1 & M3-M2 & M4-M3 & Post Covid-M4 \\\hline
AI1 & 0.056 & 0.0512 & 0.0511 & 0.105 & 0.435 \\ \hline 
AI2 & 0.048 & 0.031 & 0.020 & 0.070 & 0.505 \\ \hline 
AI3 & 0.078 & 0.080 & 0.110 & 0.183 & 0.465

 \\ \hline\end{tabular}
\caption{\label{tab:JSD table 2}JS divergence between the probability distributions of each AI at each monthly checkpoint against its preceding month}
\end{table}

The temporal divergence data, as indicated by the Jensen-Shannon Divergence (JSD) values, provides insights into the consistency of the AI models over time. 

For AI1, the JSD values relative to the reference data start at 0.056, demonstrating a close match with historical performance. The values remain relatively stable in the initial months (0.0512 in M2-M1 and 0.0511 in M3-M2), suggesting consistent performance. However, there is a noticeable increase to 0.105 in M4-M3, crossing the tolerance range, which indicates a significant shift in the model's predictive behavior that may necessitate review. The post-COVID period shows a substantial rise to 0.435, suggesting a marked deviation from the model's historical performance likely influenced by the pandemic's impact on clinical presentations.

AI2 maintains even lower initial JSD values (0.048 in M1-Reference Data), which decrease further in subsequent months, reflecting very stable performance. This trend shifts in M4-M3, where the value rises to 0.070, still within the tolerance range but denoting a potential trend towards variability that requires attention. The Post-COVID JSD value escalates dramatically to 0.505, signifying a profound divergence from previous patterns, potentially due to the evolving clinical landscape during the pandemic.

AI3 starts with a higher JSD value of 0.078 in M1-Reference Data, and it shows an incremental increase in the following months, with the M3-M2 reaching 0.110. This trend of increasing JSD continues, culminating in 0.183 in M4-M3, suggesting a continuous drift from its historical prediction patterns. The value escalates further to 0.465 in the post-COVID period, which is well above the tolerance threshold and highlights a considerable change likely exacerbated by the pandemic conditions.

In summary, all three models exhibit a significant divergence in the post-COVID era compared to previous months, with JSD values soaring beyond the 20 percent tolerance range. This indicates a substantial impact of the pandemic on the models' predictive behavior, underscoring the need for potential re-calibration or adaptation of the models to the new clinical data patterns.

\section{Discussion}

The study accentuates the vitality of regular monitoring and performance evaluation for AI models in clinical settings. Variations in predictions, especially in unprecedented situations like the onset of COVID-19, can have significant implications. It's imperative to have a robust monitoring mechanism that not only gauges the performance but also ensures the safety and efficacy of AI models in real-world scenarios. Future work can delve deeper into understanding the specific causes of these variances and establishing potential solutions.

 \subsection{From Theory to Practice challenges:}

The transition from theoretical design to practical implementation of this monitoring system involves several important considerations. Firstly, the selection of appropriate support models is a critical step. These models should ideally have a performance profile that complements the main model, thereby allowing for a more comprehensive estimation of accuracy. 

Secondly, the choice of divergence measure can significantly impact the system's ability to accurately detect discrepancies between models. The selection of an appropriate divergence measure may therefore require some trial and error, depending on the specific characteristics of the models and data \cite{21Zeng2013}.

Thirdly, the choice of the time window for historical performance comparison \cite{22lau2017chapter10} is another crucial factor that can influence the sensitivity of the system to changes in the main model's performance. This requires careful consideration and potentially some experimentation to determine the optimal window size.

Fourthly, the cost-effectiveness of the monitoring system should also be evaluated \cite{4Feng2022, 23Vanem2010}. Given that the main model and the support models are third-party solutions billed per inference, cost can be a significant factor. Optimizing the system to minimize the number of unnecessary inferences while ensuring accurate performance monitoring can be a complex balancing act.

Finally, the effectiveness of the monitoring system in detecting model decay should be evaluated regularly \cite{4Feng2022, 11soin}. This involves monitoring the divergence measures and the model's historical performance in real-time, and taking corrective action if signs of model decay are detected.

\section{Conclusion}
The increasing integration of AI models in clinical practice necessitates robust, real-time performance monitoring systems. Our proposed approach, centered around divergence measures and historical performance comparison, offers a promising solution to this challenge. By facilitating accurate estimation of model accuracy and precision in the absence of ground truth data, our system can help ensure the reliable application of AI models in real-world clinical settings. While our retrospective analysis provides a practical demonstration of this approach, ongoing research, and development are required to optimize and adapt this system to various clinical contexts.

By combining these two novel metrics, we can simultaneously monitor the main model's estimated accuracy (based on the support models' predictions) and consistency (based on the model's historic performance). This approach allows for real-time detection of potential issues with the AI model's performance, even in the absence of real-time ground truth data

\bibliographystyle{unsrt}
\bibliography{reference}

\end{document}